\pdfoutput=1

\documentclass[11pt]{article}
\usepackage[]{acl}

\usepackage[T1]{fontenc}

\usepackage{times}
\usepackage{url}
\usepackage{latexsym}
\usepackage{enumitem}
\usepackage{color, colortbl}
\usepackage{threeparttable,booktabs}
\usepackage{amssymb}
\usepackage{graphicx,subcaption}

\usepackage[hang,flushmargin]{footmisc}

\usepackage{microtype}

\usepackage{booktabs}
\usepackage{multirow}
\usepackage{adjustbox}
\usepackage{arydshln}
\usepackage{footnote}
\usepackage{tabularx}
\usepackage{array}
\usepackage{graphicx}
\usepackage{enumitem}
\usepackage{amsmath}
\usepackage{amsfonts}
\usepackage{dsfont}
\usepackage{pifont}
\usepackage{xspace}

\newcommand{\tinyflameemoji}[0]{\includegraphics[height=.012\textwidth]{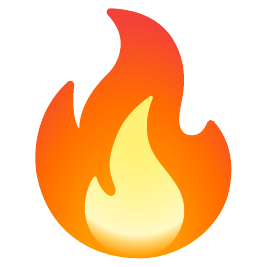}}
\newcommand{\flameemoji}[0]{\includegraphics[height=.02\textwidth]{flame_emoji.pdf}}
\newcommand{\pt}{$\mathrm{PT}$}
\newcommand{\sft}{$\mathrm{SFT}$}
\newcommand{\dpo}{$\mathrm{DPO}$}
\newcommand{\ragpt}{$\mathrm{PT}^{\text{\tiny RAG}}$}

\newcommand{\factonlysft}{$\mathrm{SFT}^{\textrm{fact}}$}
\newcommand{\fasft}{$\mathrm{SFT}^{\text{\tinyflameemoji}}$}
\newcommand{\fadpo}{$\mathrm{DPO}^{\text{\tinyflameemoji}}$}
\newcommand{\factonlydpo}{$\mathrm{DPO}^{\textrm{fact}}$}
\newcommand{\selfrm}{$\mathrm{RM}^{\text{\tiny IF}}$}
\newcommand{\factrm}{$\mathrm{RM}^{\text{fact}}$}
\newcommand{\human}{$\mathrm{Human}$}
\newcommand{\flame}{\textsc{Flame}}

\definecolor{bluencs}{rgb}{0.0, 0.53, 0.74}
\newcommand{\xmark}{\ding{55}}%

\definecolor{Gray}{gray}{0.9}

\title{\flame\textsuperscript{\flameemoji}: Factuality-Aware Alignment for Large Language Models}

\author{Sheng-Chieh Lin$^1$\thanks{\ \ This work is done during Sheng-Chieh's internship at Meta.}\ , 
Luyu Gao$^2$,
Barlas Oguz$^3$,\\
{\bf Wenhan Xiong$^3$},
{\bf Jimmy Lin$^1$},
{\bf Wen-tau Yih$^3$}, \and
{\bf Xilun Chen$^3$\thanks{\ \ Xilun and Sheng-Chieh contributed equally to this work.}}\\[1ex]
        University of Waterloo$^1$, Carnegie Mellon University$^2$, Meta AI$^3$\\[1ex]
        \texttt{s269lin@uwaterloo.ca, xilun@meta.com}
}

\begin{document}
\maketitle
\begin{abstract}
Alignment is a standard procedure to fine-tune pre-trained large language models (LLMs) to follow natural language instructions and serve as helpful AI assistants. 
We have observed, however, that the conventional alignment process fails to enhance the factual accuracy of LLMs, and often leads to the generation of more false facts (i.e.~\emph{hallucination}). 
In this paper, we study how to make the LLM alignment process more factual, by first identifying factors that lead to hallucination in both alignment steps:\ supervised fine-tuning (SFT) and reinforcement learning (RL).
In particular, we find that training the LLM on new knowledge or unfamiliar texts can encourage hallucination.
This makes SFT less factual as it trains on human labeled data that may be novel to the LLM.
Furthermore, reward functions used in standard RL can also encourage hallucination, because it guides the LLM to provide more helpful responses on a diverse set of instructions, often preferring longer and more detailed responses.
Based on these observations, we propose \emph{\textbf{f}actua\textbf{l}ity-aware \textbf{a}lign\textbf{me}nt} (\flame\textsuperscript{\tinyflameemoji}), comprised of \emph{factuality-aware SFT} and \emph{factuality-aware RL} through direct preference optimization. 
Experiments show that our proposed factuality-aware alignment guides LLMs to output more factual responses while maintaining instruction-following capability.

\end{abstract}

\maketitle

\section{Introduction}
Alignment is a standard procedure to make pre-trained large language models (LLMs)~\citep{gpt3, llama2} follow natural language instructions and serve as helpful AI assistants. 
Despite significant progress in instruction tuning~\citep{self-instruct, lima, li2024selfalignment} and LLM alignment~\citep{rlhf, anthropic_rlhf, selfrewarding}, state-of-the-art LLMs are still prone to generate false claims~\citep{factscore}. 
This motivates us to study the underlying causes of LLM hallucination as well as its relation to the alignment procedure. 

\begin{figure}[t]
    \centering
\includegraphics[width=\columnwidth]
{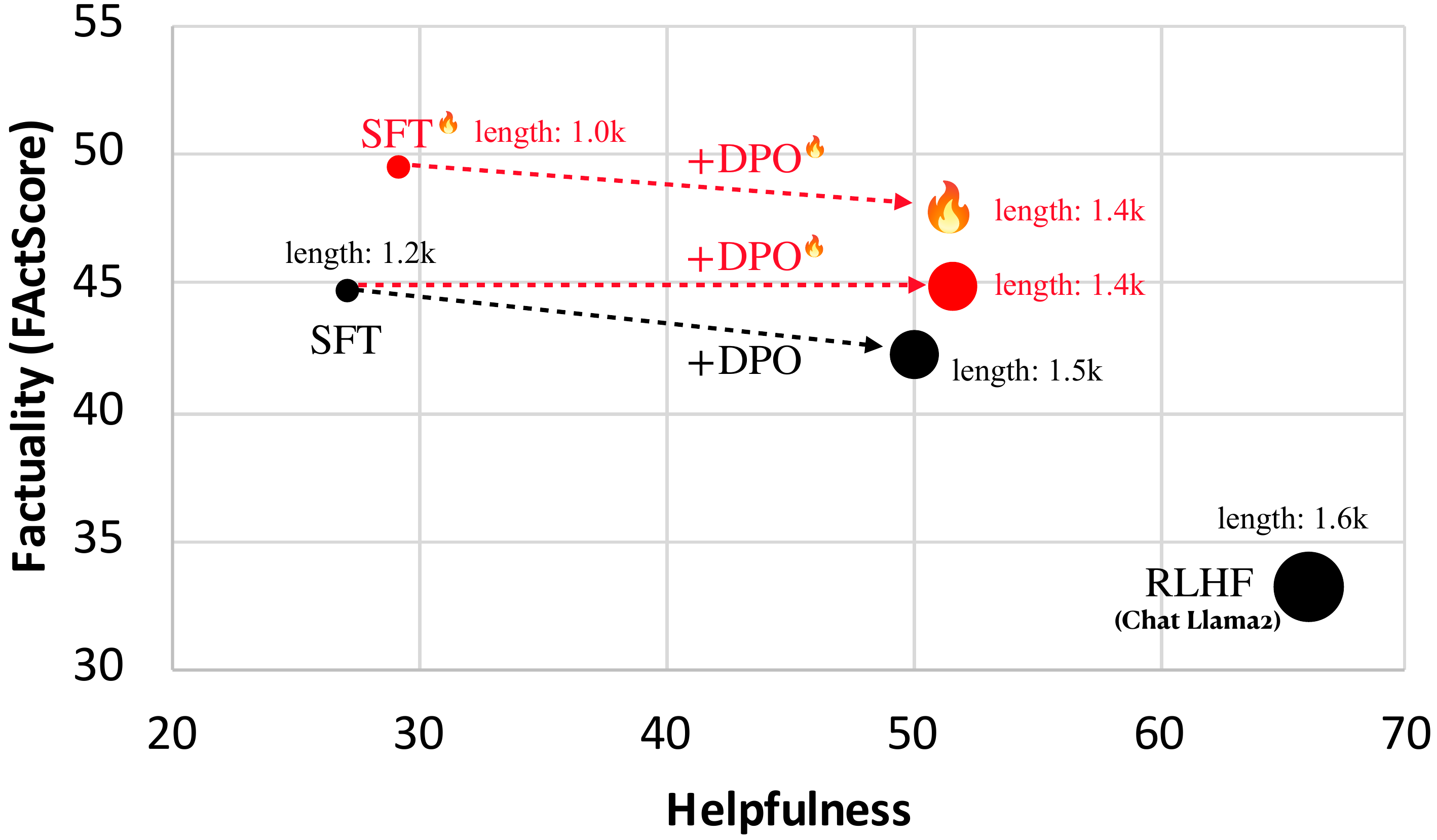}
\caption{Models' helpfulness on Alpaca Eval vs factuality on biography. Helpfulness is measured by models' win rate over our baseline \sft{} + \dpo{} on Alpaca Eval. Dot size represents averaged length of bio generation.} 
    \label{fig:teaser}
\end{figure}

The standard alignment process consists of two training phases:\ (1) supervised fine-tuning (SFT)~\citep{sanh2022multitask}; (2) reinforcement learning (RL) with human ~\citep[RLHF,][]{rlhf, anthropic_rlhf} or automated feedback~\citep[RLAIF,][]{rlaif}. 
In our study, we find that both the SFT and the RL steps in the standard alignment process may actually \emph{encourage} LLMs to hallucinate. 
First, in the SFT stage, LLMs are fine-tuned with diverse instructions paired with human created high-quality responses.
While this leads to strong instruction following capability~\citep{lima}, our study shows that such human labeled responses may present \emph{new or unknown information} to the LLM.
This, in turn, may inadvertently promote hallucination.
Second, we find that the standard reward used in the RL stage often prefers longer and more detailed responses~\citep{singhal2023long, selfrewarding}, which tends to stimulate the LLM to yield more false claims, as shown in the black dots in Figure~\ref{fig:teaser}. 
One possible reason is that most existing RLHF or RLAIF approaches rely on a single scalar reward to represent preference, which struggles to cover multiple alignment skill sets~\citep{ye2024flask} and is likely to under-present the aspect of factuality~\citep{hosking2024human}.

To address the aforementioned issues, we study the key factors which impact factuality during alignment.
In particular, we first conduct a pilot study on the biography generation task~\citep{factscore} in a more controlled setting where the alignment process focuses solely on factuality (Section~\ref{sec:pilot_study}).
Our pilot study reveals that a LLM hallucinates more if it is fine-tuned on new knowledge in either the SFT or the RL stage.
For example, a LLM becomes significantly less factual when fine-tuned on responses produced by a model with access to external knowledge (e.g.~a retrieval-augmented LLM), even though those responses are more factual themselves.
Similarly, hallucination is greatly increased if RLAIF is performed on preference pairs that consist of retrieval-augmented LLM output as positive examples and the LLM's own output as negative examples.
As a result, we discover that fine-tuning a pre-trained LLM on (a selected subset of) its own generations yields more factual responses and reduces hallucinations.

Our ultimate goal is to improve the factuality of the standard alignment process, which is challenging since LLMs may be given diverse and complex instructions.
As shown in Figure~\ref{fig:instruction_classification}, we observe that some instructions require factual responses while the others do not. 
Motivated by the observation, we propose factuality-aware alignment. 
We first identify fact-based instructions that require factual responses.
For fact-based instructions, we leverage the findings in our pilot study to create additional training data at both SFT and RL stages to explicitly guide LLMs to output factual responses.
Specifically, at the SFT stage, for fact-based instructions, instead of using human created seed training data, we construct few-shot demonstrations (from the same seed data) and generate training data using the pre-trained LLM's own knowledge. 
This can prevent fine-tuning the LLM on knowledge unknown to itself.
At the RL stage, we create additional preference pairs focused on factuality for fact-based instructions, which are combined with the standard preference pairs for instruction following during Direct Preference Optimization~\citep[DPO;][]{dpo}.

We evaluate models on Alpaca Eval~\citep{alpaca_eval} and Biography, using win rate for instruction following capability and FActScore~\citep{factscore} for factuality evaluation, respectively.
As shown in Figure~\ref{fig:teaser}, using our \flame\textsuperscript{\tinyflameemoji} method (\fasft\ + \fadpo), a significantly higher FActScore (+5.6 pts) is achieved compared to the standard alignment process (\sft{} + \dpo), without sacrificing the LLM's instruction following capability (51.2\% win rate). 
Our ablation study also indicates that identifying fact-based instructions is the key to factual alignment in the general alignment setting.

\section{Related Work}
\paragraph{Alignment.}
Since pre-trained LLMs cannot accurately follow human instructions, a bunch of work has been proposed to improve LLM alignment through SFT and RL. 
Some propose to improve SFT through data curation~\citep{lima, chen2024alpagasus}, diverse instruction augmentation~\citep{self-instruct, li2024selfalignment} while others focus on 
RL with human feedback~\citep{rlhf, anthropic_rlhf}, AI feedback~\citep{rlaif, sun2024salmon, selfrewarding}. 
The main goal of these alignment approaches is instruction following capability (or helpfulness), which may guide LLMs to output detailed and lengthy responses~\citep{singhal2023long} but inevitably encourage hallucination.   

\paragraph{Factuality.} 
Prior work has highlighted the issue of hallucination in LLMs~\citep{Kandpal2023hallucinate, mallen-etal-2023-trust}. 
To address the issue, important research lines are factuality evaluation~\citep{factscore, Wang2023FactcheckGPTEF,chern2023factool} and improvement. 
Some training-free approaches to improve LLMs' factuality include external knowledge augmentation~\citep{Kandpal2023hallucinate, cheng2023prompting,jiang-etal-2023-active} and specialized decoding~\citep{li2023inferencetime, chuang2024dola}. 

Recent studies apply RL to improve LLMs' factuality. 
For example, \citet{tian2024finetuning} propose to construct factuality preference pairs for direct preference optimization~\citep[DPO;][]{dpo}, which is closely related to our work. 
However, they focus solely on enhancing LLMs' factuality through DPO but overlook its potential impact on the models' instruction-following capability, as demonstrated in our experiments. 
In contrast, our work provides a comprehensive examination of improving LLMs' factuality and instruction-following ability through fine-tuning approaches encompassing both SFT and DPO. 
Concurrent to our work, \citet{kang2024unfamiliar} find that LLMs tend to hallucinate when facing unfamiliar queries. 
They consider improving LLMs' factuality as teaching LLMs to output abstaining or less detailed responses on such unfamiliar queries, a similar behavior observed from our LLMs fine-tuned with \flame{} (see case studies in Section~\ref{subsec:dicussions}). 
It is worth mentioning that both prior studies focus on a simplified scenario as our pilot study in Section~\ref{sec:pilot_study}: fine-tuning LLMs to improve factuality on a single task (e.g., fine-tuning and evaluating on biography generation). 
In contrast, we consider the general alignment task, where LLMs are given diverse and complex instructions.

\section{A Pilot Study on Factual Alignment}
\label{sec:pilot_study}
In this section, we first study how to align large language models (LLMs) to be more factual. 
We use biography generation as the task of our pilot study for two main reasons:\ (1) Biography generation is a simplified setting where factuality is the sole focus of the alignment process. As we will discuss in Section~\ref{sec:factuality_aware_alignment}, studying factual alignment on diverse human instructions is more complex, as the alignment process encompasses aspects beyond factuality, such as helpfulness and safety. 
(2) Evaluating the factuality of biography generation is relatively easy since Wikipedia covers sufficient information for public figures and most the facts about a person is non-debatable~\citep{factscore}.

\subsection{Alignment for Biography Generation}
A standard alignment procedure consists of supervised fine-tuning (SFT) and reinforcement learning (RL). 
In this pilot study, our main goal is to teach LLMs to generate biography with reduced misinformation. 
For the experiment, we compile training and evaluation datasets comprising 500 and 183 diverse human entities, respectively (further details provided in Appendix~\ref{appendix:biography_data_generation}). 
We employ FActScore~\citep[FS;][]{factscore} as the automated metric for assessing factuality, given its fine-grained evaluation capabilities for long-form text generation and its strong correlation with human judgments.\footnote{We use the evaluator:\ \texttt{retrieval+llama+npm}}
To study factuality alignment in this pilot study, we posit that training data is needed where the responses are more factual than the LLM's own generations. 
Thus, we use retrieval-augmented LLMs~\citep[RAG;][]{rag} to generate training data, which has been shown to output more factual responses~\citep{mialon2023augmented}.

Throughout the paper, we refer to the pre-trained (PT), supervised fine-tuned (SFT), and direct preference optimization (DPO) fine-tuned LLMs as \pt, \sft, and \dpo, respectively.\footnote{Note that in our experiments, we use DPO as the substitute of RL~\citep{ppo}.}

\begin{table}[t!]
	\caption{Pilot study on biography generation. Pos. denotes the positives for SFT or DPO. Neg. denotes the negatives for DPO. FS denotes FActScore.}
	\label{tb:pilot_study_bio_gen}
	\centering
	 \resizebox{1\columnwidth}{!}{  
   \begin{threeparttable}
    \begin{tabular}{l|cc|cc}
\hline \hline
\multirow{2}{*}{Llama-2 7B}& \multicolumn{2}{c|}{src. of supervision}& \multicolumn{2}{c}{\textbf{Bio}} \\
\cmidrule(rl){2-3} \cmidrule(rl){4-5} 
 & Pos.& Neg.& FS& \# Corr. / Err.\\
 \hline
 (1) \pt& -& -&	39.1&	14.4 / 22.0\\
 (2) \ragpt& -& -&55.4&	18.6 / 15.9\\
 \hline
(3) \multirow{2}{*}{\sft} & \pt& -&	37.9&	13.4 / 21.8\\
(4) & \ragpt& -&	35.7&	13.5 / 23.7\\
 \hline
(5) \multirow{2}{*}{\dpo} & \pt\tnote{$\ast$}& \pt\tnote{$\ast$}&	41.6&	15.4 / 20.7\\
(6) & \ragpt& \pt&	23.5& 12.7 / 34.9\\
\arrayrulecolor{black}
\hline \hline
	\end{tabular}
 		\begin{tablenotes}
	\item[$\ast$] FActScore is used to select positives and negatives. 
    \end{tablenotes}
    \end{threeparttable}
	}
 \vspace{-0.4cm}
\end{table}

\paragraph{SFT.} 
We explore two sources of supervision to generate training data (detailed in Appendix~\ref{appendix:biography_data_generation}):\ (1) using \ragpt{} with few-shot demonstration to generate biographies for each name entity in training data, where \ragpt{} is \pt{} augmented with an off-the-shelf retriever~\citep{dragon}; (2) using vanilla \pt{} with few-shot demonstration to generate training data as a baseline. 
As shown in Table~\ref{tb:pilot_study_bio_gen}, \ragpt{} is indeed much more factual than \pt. 
However, a surprising discovery in the pilot study is that \emph{fine-tuning on such more factual instruction--biography pairs generated by \ragpt{} results in a less factual \sft{} model} (row 4 vs 3).  

\paragraph{DPO.} 
We further fine-tune the LLMs to be more factual through DPO. 
An intuitive way to create factuality preference pairs is to directly use the samples from \ragpt{} and \pt{} as positives and negatives since \ragpt{} generates more factual biographies than \pt{} (row 2 vs 1). 
Another approach is to employ FActScore (FS) as the reward to select positive and negative samples among the generations from \pt{} itself~\citep{tian2024finetuning} (detailed in Apppendix~\ref{appendix:biography_data_generation}).
As shown in Table~\ref{tb:pilot_study_bio_gen}, \dpo{} fine-tuned on self-generated data with FS reward guides models to generate more factual responses (row 5 vs 3); however, \dpo{} fine-tuned with the supervision of \ragpt{} makes the models hallucinate even more than its \sft{} counterpart (6 vs 4). 

This outcome suggests that compelling models to generate responses akin to \ragpt{} prompts increases hallucination. 
Conversely, fine-tuning LLMs on their own generations appears to be crucial for factual alignment, a finding applicable to both SFT and DPO fine-tuning.

\subsection{Strategies for Factual Alignment}
\label{subsec:strategy_for_factual_alignment}
From the pilot study, we find that better quality data (in terms of factuality) for SFT and DPO does not necessarily yield models with better factual alignment. 
This is likely because the supervision from RAG contains information unknown to the LLM; thus, fine-tuning on RAG generated responses may inadvertently encourage the LLM to output unfamiliar information. 
To avoid unknown knowledge from being presented to the LLM, a viable strategy is to create SFT and DPO training data using the generated responses from the LLM itself.

\begin{figure}[t]
    \centering
\includegraphics[width=\columnwidth]
{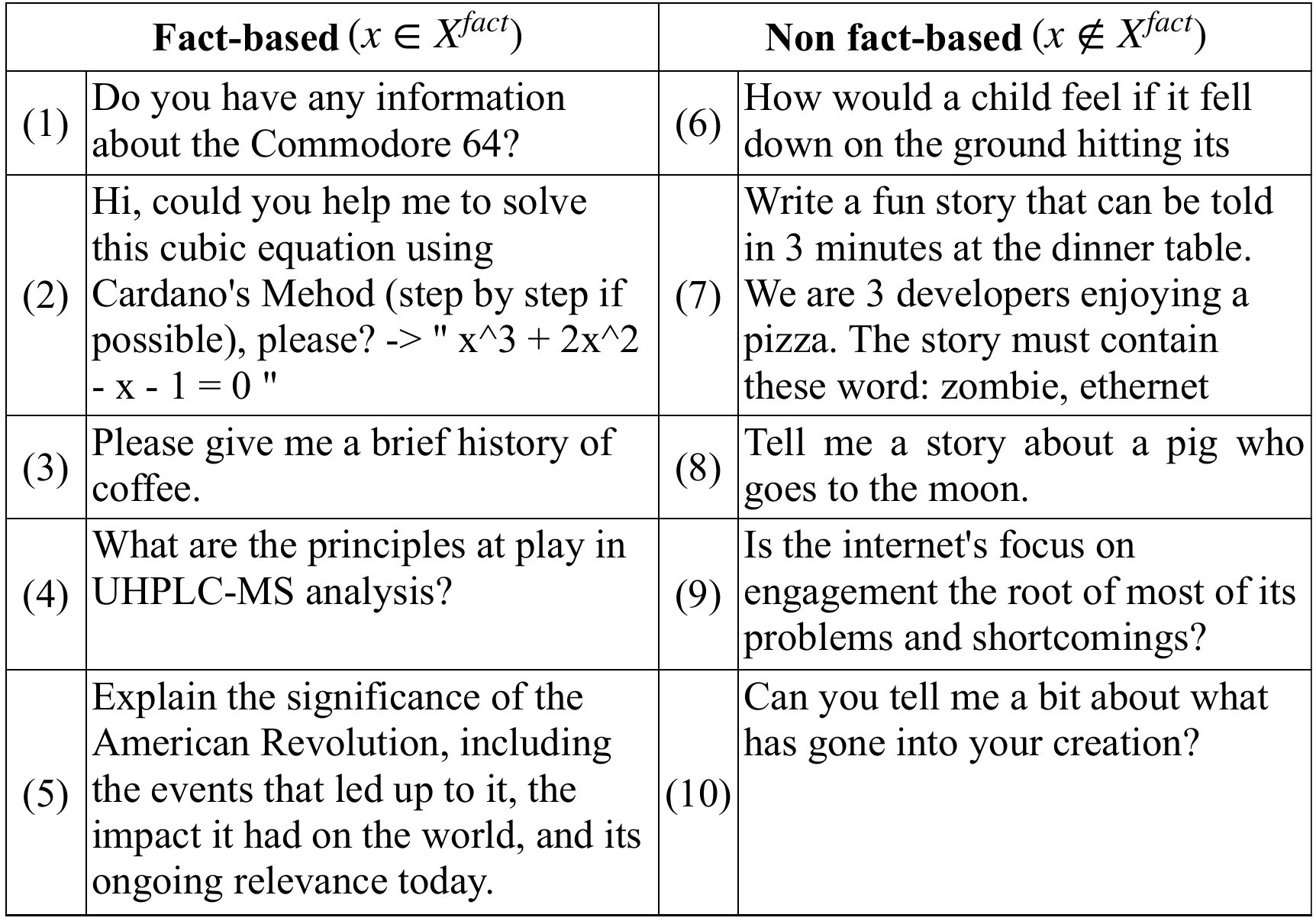}
\caption{Instructions from Open Assistant dataset. The instructions are classified with \sft{} model using the prompt in Appendix, Figure~\ref{fig:prompt_for_instruction_classifier}.} 
    \label{fig:instruction_classification}
\end{figure}

\section{Factuality-Aware Alignment}
\label{sec:factuality_aware_alignment}
In the section, we further extend our discussion of factual alignment to encompass more general instructions. 
Unlike biography generation in Section~\ref{sec:pilot_study}, where factuality is the main alignment objective, human instructions are diverse and complex, necessitating a range of alignment skill sets beyond factuality alone; e.g., logical thinking, problem handling and user alignment~\citep{ye2024flask}. 
Thus, conducting factual alignment with the diverse instructions face two main challenges:\ (1) different instructions may demand distinct skill sets. 
For example, in Figure~\ref{fig:instruction_classification}, instruction 3, ``Please give me a brief history of coffee'', necessitates factual accuracy and concise summarization, while instruction 8, ``Tell me a story about a pig who goes to the moon'', prioritizes creativity and imagination over strict factuality. 
(2) As recent studies have emphasized~~\citep{ye2024flask,hosking2024human}, using a single scalar for reward modeling fails to adequately address multiple alignment skill sets and often under-presents the aspect of factuality. 

To tackle the aforementioned challenges, we propose \emph{\textbf{f}actua\textbf{l}ity-aware \textbf{a}lign\textbf{me}nt} (\flame\textsuperscript{\tinyflameemoji}). 
To address the first challenge, we propose to prompt LLMs to classify whether a given instruction demands the response to be factual, as shown in Figure~\ref{fig:instruction_classification}. 
We then apply the factuality fine-tuning strategy for SFT and DPO discussed in Section~\ref{subsec:strategy_for_factual_alignment} to those fact-based instructions. 
Furthermore, to address the second challenge, we employ separate rewards to evaluate the factuality and instruction following capability of a LLM.  
For simplicity, our work only considers two alignment skill sets:\ instruction following and factuality. 
We leave more comprehensive reward modeling to future work.

In the following, we first describe our baseline alignment approach and introduce our proposed factuality-aware alignment built on top of the baseline alignment procedure.

\subsection{Baseline Alignment}
\label{subsec:baseline_alignment_trainin}
We initialize \pt{} from Llama-2 70B pre-trained model\footnote{\href{https://huggingface.co/meta-llama/Llama-2-70b}{meta-llama/Llama-2-70b}} and build our baseline alignment procedure following self-rewarding language models~\citep{selfrewarding} due to its simplicity and independence of other strong LLMs (e.g., GPT4) or human evaluators as a reward model. 
The alignment comprises two steps: (1) building \sft{} model fine-tuned on a high-quality seed data consisting of 3,200 instructions and each instruction is paired with the best response created by humans from Open Assistant dataset~\citep[OASST;][]{oasst}; (2) further fine-tuning \sft{} through DPO on instruction following preference data $(x, y_{+}, y_{-})$ constructed by itself (\sft) as the reward model, \selfrm, where $y_{+}$ and $y_{-}$ are the positive and negative responses for a given prompt $x$, respectively. 
The resulting fine-tuned model is denoted as \sft{} + \dpo. 
Note that, following \citet{selfrewarding}, we use additional augmented 20K instructions to create the preference training data for DPO fine-tuning. 
Further details are provided in Appendix~\ref{appendix:alignment_with_self_rewarding}.

\begin{figure}[t]
    \centering
    \resizebox{1\columnwidth}{!}{
\includegraphics{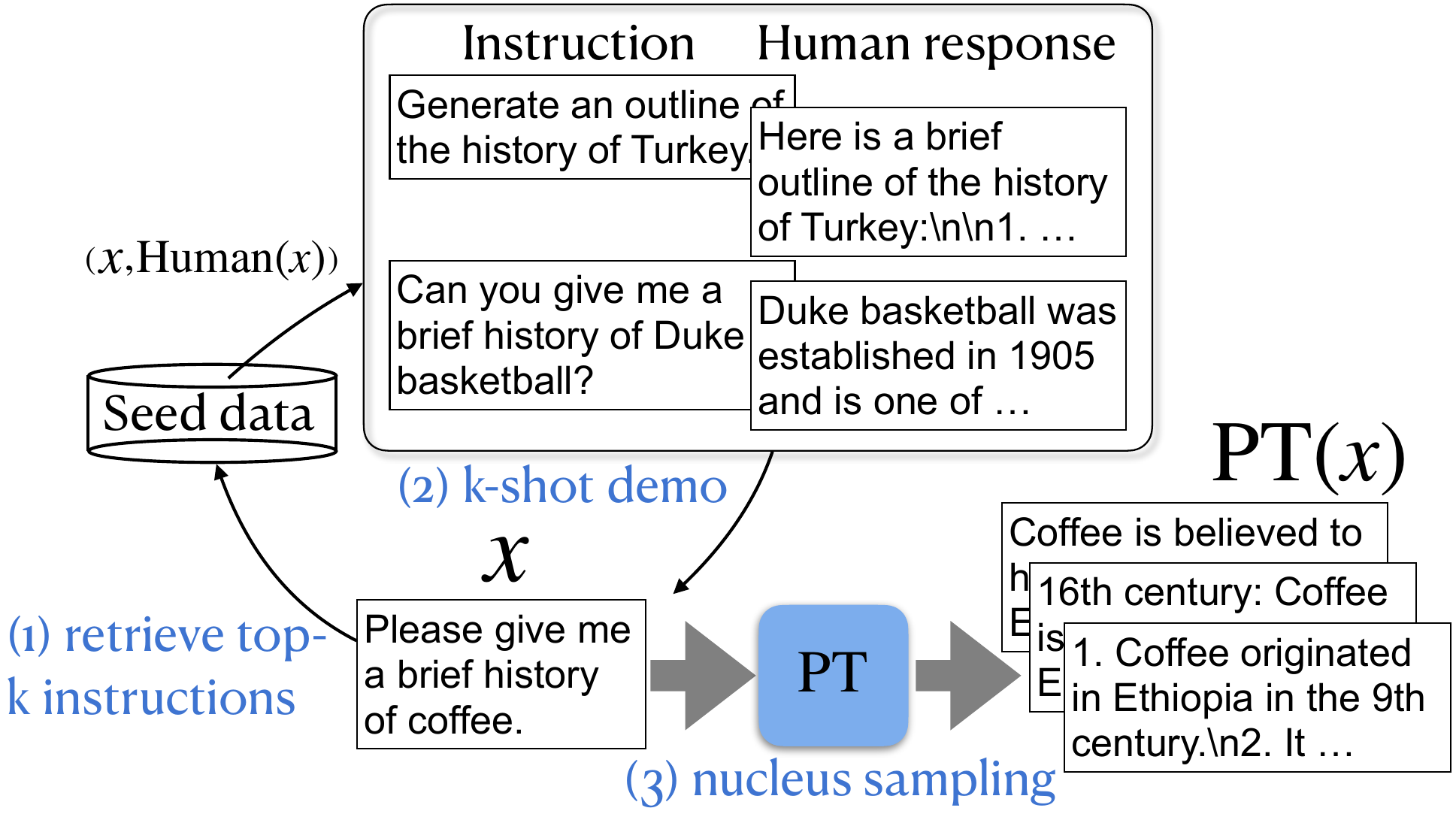}
    }
    \caption{Illustration of response generation using a pre-trained LLM (\pt) with few-shot demonstration.} 
    \label{fig:fewshot_gen_with_pt}
\end{figure}

\subsection{Our Approach}
\subsubsection{Factuality-Aware SFT (\fasft)} 
Although leveraging human created high-quality seed data is a reasonable choice for SFT~\citep{lima}, our study in Section~\ref{sec:pilot_study} suggests that fine-tuning on such high-quality data generated by models other than the LLM itself may present unknown information to the LLM, which may in turn encourage hallucination. 
To address the above issue, for each instruction from the seed data, we elicit the knowledge from the pre-trained LM itself by generating the responses with few-shot demonstration. 
Furthermore, to better use the knowledge from both humans and the pre-trained LLM itself, we propose to utilize human generated responses for non-fact-based instructions, while leveraging the responses sampled from pre-trained LLMs for fact-based instructions to mitigate the introduction of unknown knowledge.

Specifically, we create factuality-aware alignment training data for SFT with two steps. 
(1) Classifying instructions:\ we first prompt \sft{} to judge whether an instruction from the seed data is fact-based ($x \in X^{\text{fact}}$) or not.\footnote{Prompt for fact-based instruction classification is shown in Appendix, Figure~\ref{fig:prompt_for_instruction_classifier}.} 
(2) Eliciting knowledge from \pt{}:\ as illustrated in Figure~\ref{fig:fewshot_gen_with_pt}, we sample 10 responses from \pt{} with 5-shot demonstration, $(x_0, \text{\human}(x_0)) \cdots(x_4, \text{\human}(x_4))$, where $x_k$ is the top-$k$ similar instruction to $x$ retrieved by DRAGON+~\citep{dragon} from the seed data.
$\text{\human}(x_k)$ denotes the corresponding human response to $x_k$ in the seed data.

\begin{figure}[t]
    \centering
    \resizebox{1\columnwidth}{!}{
\includegraphics{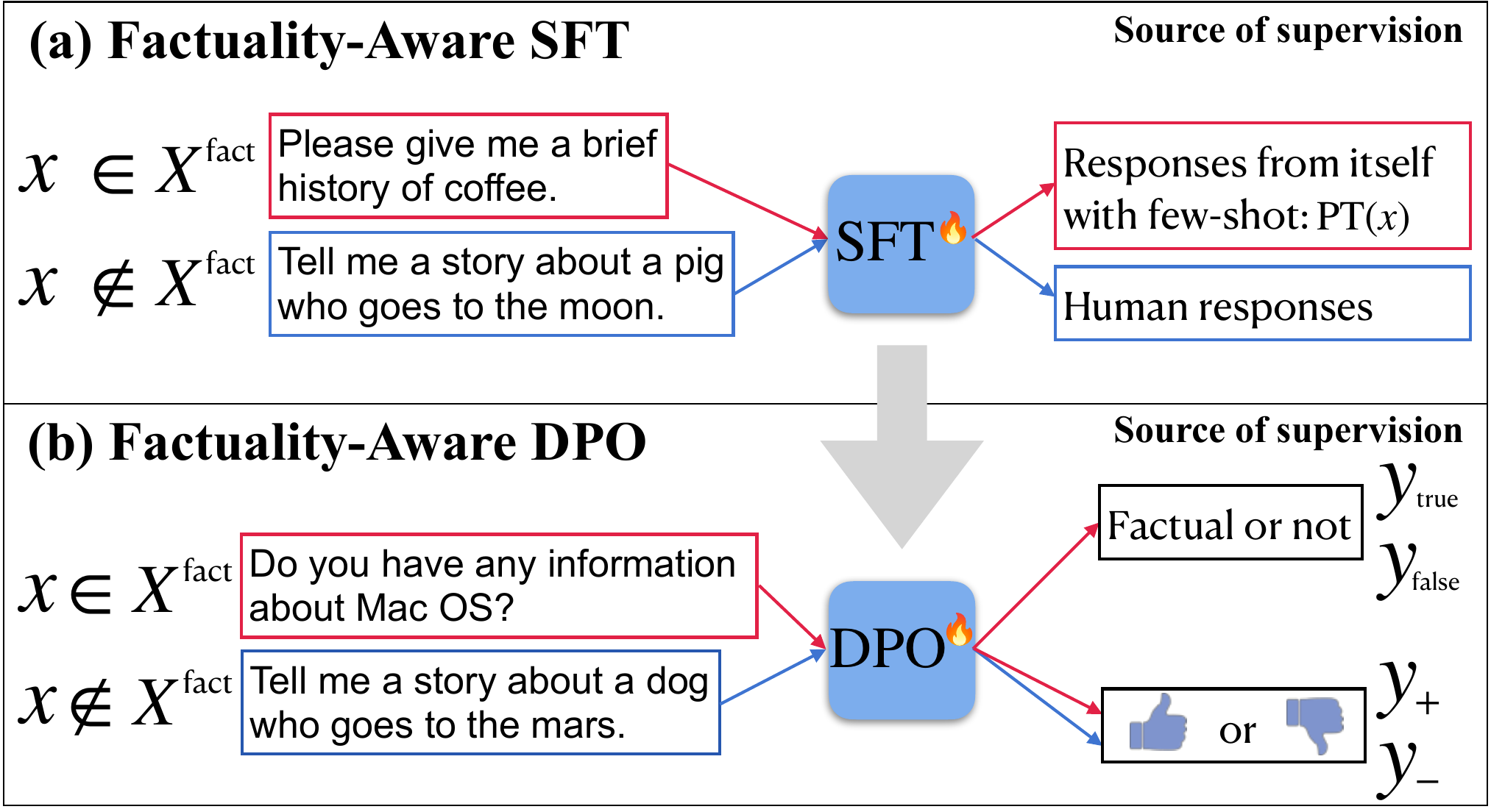}
    }
    \caption{Illustration of factuality-aware alignment.} 
    \label{fig:fa_alignment}
\end{figure}

As illustrated in Figure~\ref{fig:fa_alignment}(a), the resulting training data for SFT is $(x\notin X^{\text{fact}}, \text{\human}(x)), (x\in X^{\text{fact}}, \text{\pt}(x))$, where \pt($x$) denotes the set of responses to $x$ sampled from \pt{}. 
The resulting fine-tuned model is denoted as \fasft{}.

\subsubsection{Factuality-Aware DPO (\fadpo)} 
\label{subsubsec:factuality_aware_dpo}
At the second stage of alignment with DPO, we use \fasft{} to generate multiple responses $y_0, y_1, \cdots$ for a given instruction $x$; then, using \fasft{} itself as the reward model (\selfrm) to create a preference pair: $(x, y_+, y_{-})$.\footnote{We sample 4 responses for each augmented instruction.} 
The above data creation procedure is the same as the second stage of our baseline alignment in Section~\ref{subsec:baseline_alignment_trainin}. 
However, recent studies~\citep{saha2023branchsolvemerge,hosking2024human,ye2024flask} indicate that a single scalar reward from human feedback or LLM reward models may under-represents the aspect of factuality. 
To address this limitation, we introduce another factuality reward model (\factrm) to evaluate factuality of responses and create a factuality preference pair for fact-based instructions:\ $(x \in X^{\text{fact}}, y_{\text{\tiny true}}, y_{\text{\tiny false}})$.

Specifically, we build \factrm{} with retrieval augmentation to measure the percentage of facts in a response that are correct. \factrm{} comprises two main components:\ atomic fact decomposition and retrieval augmented claim verification. 
We detail the components and ablate their impacts on the quality of \factrm{} in Appendix~\ref{appendix:factuality_reward_modeling}.  
We compute factuality reward for the same responses sampled from \fasft{}:\ $\text{\factrm}(x, y_0),\text{\factrm}(x, y_1), \cdots$. 
The response with the highest (lowest) factuality reward is chosen as $y_{\text{\tiny true}}$ ($y_{\text{\tiny false}}$). 
Note that if the chosen paired responses show large difference in instruction-following reward, we discard the pair; i.e., $|\text{\selfrm}(x, y_{\text{\tiny true}}) - \text{\selfrm}(x, y_{\text{\tiny false}}) | > 0.5$. 
As illustrated in Figure~\ref{fig:fa_alignment}(b), in factuality-aware DPO training, the model is initialized from \fasft{} and the fine-tuned model is our final factuality-aware aligned model, denoted \fasft{} + \fadpo. 
The specific procedures for fine-tuning models in both the SFT and DPO are described in Appendix~\ref{appendix:training_details}.

\begin{table*}[t!]
	\caption{Experimental results of supervised fine-tuning on Open Assistant dataset. \pt{} denotes pre-trained Llama2 70B with 5-shot demonstration. \factonlysft{} denotes the variant which only optimizes factuality. FS denotes FActScore.}
	\label{tb:main_sft_results}
	\centering
	 \resizebox{1\textwidth}{!}{  
  \begin{threeparttable}
    \begin{tabular}{l|cc|c|cc|cc|cc}
\hline \hline
\multirow{2}{*}{Llama-2 70B}& \multicolumn{2}{c}{src. of supervision}& \multicolumn{1}{c}{\textbf{Alpaca Eval}}& \multicolumn{2}{c}{\textbf{\textbf{Bio}}}&\multicolumn{2}{c}{\textbf{Alpaca Fact}}& \multicolumn{2}{c}{\textbf{FAVA}}\\
\cmidrule(rl){2-3} \cmidrule(rl){4-4} \cmidrule(rl){5-6} \cmidrule(rl){7-8} \cmidrule(rl){9-10}      
& \human& \pt& win rate over (1)& FS& \# Corr. / Err.& FS& \# Corr. / Err.& FS& \# Corr. / Err.\\
 \hline
 (0) \pt& -& -& -& 53.1& 15.3 / 13.5& -& -& -& -\\
 \hline
 (1) \sft& \checkmark& \xmark& 50.0& 44.7& 21.1 / 26.8& 38.6& 16.7 / 29.0& \textbf{54.4}& 21.2 / 25.8\\
 (2) \factonlysft& \xmark& \checkmark& 48.1& 48.5& 19.6 / 20.6& \textbf{42.0}& 17.5 / 28.4& 53.3& 18.3 / 24.2\\
 (3) \fasft& \checkmark\tnote{$\ast$}& \checkmark\tnote{$\ast$}& \textbf{51.2}& \textbf{49.5}& 19.9 / 19.5& 41.4& 18.3 / 27.7& 54.2& 19.3 / 22.4\\

\arrayrulecolor{black}
\hline \hline
	\end{tabular}
\begin{tablenotes}
	\item[$\ast$] \fasft{} uses supervision from \human{} and \pt{} for non-fact-based and fact-based instructions, respectively. 
    \end{tablenotes}
    \end{threeparttable}
	}
\end{table*}

\section{Experiments}
\subsection{Evaluation Datasets and Metrics}
\paragraph{Instruction Following.} 
We use the the 805 instruction following tasks from Alpaca Eval~\citep{alpaca_eval} to evaluate models head to head win rate against our baselines using the recommended evaluator:\ \texttt{alpaca\_eval\_gpt4\_turbo\_fn}. 
We use \sft{} and \sft{} + \dpo{} described in Section~\ref{subsec:baseline_alignment_trainin} as the baselines for win rate comparisons.

\paragraph{Factuality.} 
We evaluate models on three datasets with diverse knowledge-intensive instructions for factuality. 
(1) Biography:\ a knowledge insensitive sub-task of instruction following tasks. 
Following our pilot study in Section~\ref{sec:pilot_study}, we use the 183 human entities provided by \citet{factscore} with the prompt ``Tell me a bio of \texttt{entity name}''. 
(2) Alpaca Fact: we extract the fact-based instructions from the 803 instructions using our \sft{} model (with the prompt shown in Appendix, Figure~\ref{fig:prompt_for_instruction_classifier}), resulting in 241 instructions.
(3) FAVA~\citep{fava}\footnote{\href{https://huggingface.co/datasets/fava-uw/fava-data/blob/main/annotations.json}{FAVA dataset}}:\ the 141 knowledge-intensive instructions from multiple sources, including Open Assistant~\citep{oasst}, No Robots~\citep{no_robots}, WebNLG~\citep{webnlg} and manually created datasets. 
We report FActScore (FS) without length penalty as the metric for all the three datasets.
Note that original FS computes proportion of correct facts with additional penalty on short generations with less than 10 atomic facts. 
This penalty aims to address situations where models provide insufficiently detailed answers. 
We assume this aspect is considered in the evaluation of instruction following in Alpaca Eval. 
In addition, we also report the number of correct and erroneous facts. 
All the numbers reported are averaged over the instructions in each dataset.

In addition, we also evaluate our fine-tuned models' truthfulness using TruthfulQA~\citep{truthfulqa}. 
We evaluate model performance in the generation task and use ROUGE~\citep{lin-2004-rouge} and BLEU~\citep{papineni-etal-2002-bleu} to measure the quality of responses.

\begin{table*}[t!]
\caption{Experiments of direct preference optimization (DPO). IF.~and Fact.~denote instruction following $(x, y_{+}, y_{-})$ and factuality $(x \in X^{\text{\tiny fact}}, y_{\text{\tiny true}}, y_{\text{\tiny false}})$ preference data, where $X^{\text{\tiny fact}}$ denotes the set of fact-based instructions. \factonlydpo{} denotes the variant which only optimizes factuality. The preference data statistics is listed in Appendix, Table~\ref{tb:training_data_statistics}.}
	\label{tb:main_dpo_results}
	\centering
	 \resizebox{1\textwidth}{!}{  
    \begin{tabular}{l|cc|c|cc|cc|cc}
\hline \hline
\multirow{2}{*}{Llama-2 70B} & \multicolumn{2}{c|}{src. of supervision}& \multicolumn{1}{c}{\textbf{Alpaca Eval}}& \multicolumn{2}{c}{\textbf{Bio}}& \multicolumn{2}{c}{\textbf{Alpaca Fact}}& \multicolumn{2}{c}{\textbf{FAVA}}\\
 \cmidrule(rl){2-3} \cmidrule(rl){4-4} \cmidrule(rl){5-6} \cmidrule(rl){7-8} \cmidrule(rl){9-10}
 & IF.& Fact.& win rate over  (2)& FS& \# Corr. / Err.& FS& \# Corr. / Err.& FS& \# Corr. / Err.\\
 \hline
 (0) Chat& \multicolumn{2}{c|}{Proprietary data}& 66.2& 33.2& 23.4 / 43.6& 39.3& 22.3 / 36.4& 47.5& 28.0 / 31.3\\
 \hline
 (1) \sft& -& -& 27.1& 44.7& 21.1 / 26.8& 38.6& 16.7 / 29.0& 54.4& 21.2 / 25.8\\
 \arrayrulecolor{lightgray}
 \hline
 \arrayrulecolor{black}
(2) + \dpo& \checkmark& \xmark& 50.0& 42.3& 24.6 / 35.0& 41.6& 22.9 / 34.6& 52.9& 28.1 / 26.8 \\
(3) + \factonlydpo& \xmark& \checkmark& 40.8& 47.1& 19.8 / 23.9& 48.2& 17.5 / 19.0& 57.9& 20.0 / 15.9 \\
(4) + \fadpo& \checkmark& \checkmark& \textbf{51.7}& 44.9& 23.7 / 30.3& 45.0& 23.1 / 28.7& 56.4 &27.1 / 23.3\\
\hline
 (5) \fasft& -& -& 29.1& \textbf{49.5}& 19.9 / 19.5& 41.4& 18.3 / 27.7& 54.2& 19.3 / 22.4\\
 \arrayrulecolor{lightgray}
 \hline
 \arrayrulecolor{black}
(6) + \dpo&\checkmark & \xmark& 50.4& 46.3& 24.0 /  28.7& 43.9& 21.6 / 28.8& 55.0& 25.4 / 22.0 \\
(7) + \fadpo& \checkmark& \checkmark& 51.2& 47.9& 25.9 / 28.5& \textbf{48.7}& 24.1 / 25.5& \textbf{58.9}& 29.0 / 22.2\\

\hline \hline
	\end{tabular}
	}
\end{table*}

\subsection{Comparisons of SFT} 
Table~\ref{tb:main_sft_results} compares the pre-trained Llama-2 70B fine-tuned on OASST dataset with responses from different sources. 
We list the FActScore (FS) of biography generation using the pre-trained model through Bio 5-shot demonstration as reference (row 0) and \sft, which is fine-tuned on our seed data with human created responses, is our baseline (row 1). 
We first notice that \sft{} shows significant FActScore degrade (53.1 vs 44.7) compared to Bio 5-shot with the pre-trained model. 
It seems that \sft{} tends to generate more lengthy responses but with more erroneous facts.

When eliciting the knowledge from \pt{} by fine-tuning on its own generated responses, \factonlysft{} generates more factual responses in Biography and Alpaca (row 2 vs 1).  
However, it shows slightly inferior instruction following capability in Alpaca Eval. 
This result demonstrates that human responses indeed teach LLMs how to better follow instructions but also encourage LLMs to output more false facts. 
On the other hand, eliciting the knowledge from the pre-trained model itself avoids the encouragement of hallucination albeit with a slight reduction in instruction-following capability. 
Finally, \fasft\, combining supervision from humans and \pt, shows comparable instruction following capability and output more factual responses on fact-based instructions (row 3 vs 1).

\subsection{Comparisons of DPO}
Table~\ref{tb:main_dpo_results} compares different DPO training recipes. 
First, we conduct DPO fine-tuning on our SFT baseline, \sft. 
When further aligning the model to follow instructions, \dpo{} sees a significant improvement in instruction following capability (row 2 vs 1) with win rate 72.9 over \sft; however, the instruction aligned model tends to output lengthy responses with more factual errors (see examples in Appendix, Figure~\ref{fig:case_studies}). 
On the other hand, when only aligned with factual preference data, \factonlydpo{} shows less improvement in instruction following capability (row 1 vs 3). 
These results indicate that preference optimization for either instruction following or factuality alone may come at the expense of the other since the former encourages models to output long and detailed responses while the later discourages models to output false claims. 
When jointly conducting instruction and factuality alignment, \fadpo{} not only better follows instructions but also outputs more factual responses (row 4 vs 1, 2). 
Finally, initializing from \fasft, the DPO fine-tuned models are more factual than their counterparts (i.e., 6 vs 2 and 7 vs 4) without instruction following capability degrade.
We also list the results from Llama-2-Chat 70B (row 0) and observe that despite of its strong instruction following capability, it tends to output many more incorrect facts. 
This results demonstrate that standard alignment, even on proprietary commercial data, may encourage LLMs to hallucinate.
In contrast, our factuality-aware alignment guides LLMs to output more factual responses without degradation in their general instruction following capabilities. 
It is worth noting that \factonlysft{} and \factonlydpo{} are similar to SFT and DPO fine-tuning proposed by \citet{tian2024finetuning}, which improve LLMs' factuality but degrade instruction following capability.

\begin{table}[t!]
	\caption{Results on TruthfulQA.}
	\label{tb:truthfulqa}
	\centering
	 \resizebox{0.95\columnwidth}{!}{  
	  \setlength\tabcolsep{0.14cm}
    \begin{tabular}{l|cc|cc}
\hline \hline
 \multirow{2}{*}{Llama-2 70B} & \multicolumn{2}{c|}{src. of supervision}& \multicolumn{2}{c}{\textbf{TruthfulQA}}\\
  \cmidrule(rl){2-3}  \cmidrule(rl){4-5}
 & IF.& Fact.& BLUE & ROUGE \\
 \hline
 (0) Chat& \multicolumn{2}{c|}{Proprietary data}& 0.21& 1.16\\
 \hline
 (1) \sft& -& -& 0.37& 0.20\\
 \arrayrulecolor{lightgray}
 \hline
 \arrayrulecolor{black}
  (2) + \dpo& \checkmark& \xmark& 0.03& 0.54\\
 (3) + \factonlydpo& \xmark& \checkmark& 0.30& 1.12\\
 (4) + \fadpo& \checkmark& \checkmark& 0.15&	0.80\\
 \hline
 (5) \fasft& -& -& 0.39& 0.51\\
 \arrayrulecolor{lightgray}
 \hline
 \arrayrulecolor{black}
 (6) + \dpo& \checkmark& \xmark&0.07& 0.91\\
 (7) + \fadpo& \checkmark& \checkmark& 0.20& 0.96\\
\arrayrulecolor{black}
\hline \hline
	\end{tabular}
	}
\end{table}

\subsection{Results on TruthfulQA}
Table~\ref{tb:truthfulqa} compares models performance on TruthfulQA. 
Generally, we observe that our factuality-aware alignment training guides LLMs to output more truthful responses. 
For example, factuality-aware SFT improves LLMs' truthfulness (row 5 vs 1).  In addition, DPO fine-tuning on the factuality preference data guides LLMs to output more truthful responses (rows 3,4 vs 2 and 7 vs 6). 
Note that we observe that \sft{} and \dpo{} models show a reverse trend in BLUE and ROUGE. 
This is likely because \sft{} models tend to generate shorter responses than the \dpo{} ones do.

\subsection{Discussions}
\label{subsec:dicussions}
\begin{table}[h]
	\caption{Effects of fact-based classification on factuality-aware alignment.}
	\label{tb:fact_based_classification_ablations}
	\centering
	 \resizebox{1\columnwidth}{!}{ 
  \begin{threeparttable}
	  \setlength\tabcolsep{0.05cm}
    \begin{tabular}{ll|cc|c|ccc}
\hline \hline

&&  \multicolumn{2}{c}{Classifier}&\multicolumn{1}{c}{\textbf{Alpaca Eval}}& \multicolumn{2}{c}{\textbf{\textbf{Bio}}}\\
 \cmidrule(rl){3-4} \cmidrule(rl){5-5} \cmidrule(rl){6-7}
&& Inst.& Sent.& win rate& FS& \# Corr. / Err.\\
 \hline
(1)&\multirow{2}{*}{\fasft}& \xmark& -& 47.6\tnote{$\ast$}& 48.4& 20.5 / 21.4\\
(2)&&\checkmark & -&51.2\tnote{$\ast$}&49.5& 19.9 / 19.5 \\
\hline
(3)&\multirow{3}{*}{\sft\ + \fadpo}& \xmark& \xmark& 46.8\tnote{$\triangle$}& 46.8& 21.7 / 25.3\\
(4)&&\checkmark & \xmark&51.7\tnote{$\triangle$}& 45.0& 23.7 / 30.3 \\
(5)&&\checkmark & \checkmark& 51.3\tnote{$\triangle$}& 42.9& 25.5 / 36.8 \\
\arrayrulecolor{black}
\hline \hline
	\end{tabular}
 \begin{tablenotes}
	\item[$\ast$] comparing with SFT baseline, \sft. 
        \item[$\triangle$] comparing with DPO baseline, \sft{} + \dpo. 
    \end{tablenotes}
    \end{threeparttable}
	}
\end{table}
\paragraph{Effects of Fact-Based Instruction Classification.}  
In our factuality-aware alignment, we prompt \sft{} to judge whether an instruction requires a factual response and apply our factuality alignment strategy to the fact-based instruction. 
Without the instruction classification, in our factuality-aware SFT, we cannot create supervision from \human{} and \pt{} responses for respective non-fact-based and fact-based instructions. 
Instead, for each instruction, we create instruction--response pairs from 1 and 10 responses from \human{} and \pt{} as supervisions, respectively. 
Note that, during fine-tuning, for each instruction, we randomly sample instruction--response pair either created from \human{} or \pt{} with same probability. 
The SFT model shows degradation in both instruction following capability and factuality results, as shown in row 1 vs 2 of Table~\ref{tb:fact_based_classification_ablations}. 
Second, for factuality-aware DPO, without the instruction classification, we create factuality preference pairs from all instructions instead of fact-based instructions. 
The DPO fine-tuned model outputs slightly more factual responses but sacrifice instruction following capability, as shown in row 3 vs 4 of Table~\ref{tb:fact_based_classification_ablations}.

\paragraph{Effects of Fact-Based Sentence Classification.} 
In addition, we observe that not all the sentences in a response to a fact-based instruction require fact check. 
For example, given the response, ``Of course. The Commodore 64 is a 8-bit home computer that was released by Commodore International in August 1982.'', conducting fact check for the first sentence ``Of course.'' is not necessary and may make the factuality reward less accurate. 
To address this issue, we prompt \sft{} to judge whether each sentence in a response required fact check using the prompt in Appendix, Figure~\ref{fig:prompt_for_claim_classifier}. 
We only conduct fact check and compute factuality rewards for those fact-based sentences. 
However, as shown in Table~\ref{tb:fact_based_classification_ablations}, computing factuality rewards for fact-based sentences makes our factual alignment less effective (row 5 vs 4). 
This is likely because the fact-based sentence classifier is not accurate enough and brings noise into our factuality reward model (see examples in Appendix, Figure~\ref{fig:fact_based_claim_check_cases}).

\begin{table}[t]
	\caption{Ablation on factuality preference data creation.}
	\label{tb:dpo_ablations}
	\centering
	 \resizebox{1\columnwidth}{!}{  
\begin{threeparttable}
    \begin{tabular}{ccc|c|c}
\hline \hline
\multicolumn{3}{c|}{Factuality preference data}&\multicolumn{1}{c}{\textbf{Alpaca Eval}}& \multicolumn{1}{c}{\textbf{\textbf{Bio}}}\\
 \cmidrule(rl){1-3} \cmidrule(rl){4-4} \cmidrule(rl){5-5}
Reward model& Pos.,Neg.& \# pairs& win rate\tnote{$\triangle$}& FS\\
\hline
\factrm& max, min& 3,315& 51.7& 44.9\\
\factrm& enum.& 5,126& 50.7& 45.0\\
\selfrm\ + $5*$\factrm& max, min& 6,340& 50.1& 45.1 \\
\hline \hline
	\end{tabular}
 \begin{tablenotes}
        \item[$\triangle$] comparing with DPO baseline, \sft{} + \dpo. 
    \end{tablenotes}
    \end{threeparttable}
	}
\end{table}
\paragraph{Ablations on Factuality Preference Data Creation.} 
In this section, we examine different ways of creating factuality preference data for factuality-aware DPO training. 
First, for each fact-based instruction, instead of choosing the responses (among the 4 generated responses) with the maximum and minimum factuality rewards (\factrm) as the respective positive and negative samples, we enumerate all the possible response pairs and choose the response with higher (lower) \factrm{} as the positive (negative) sample from each enumerated pair. 
If the difference of \factrm{} is smaller than $0.2$, we treat them as equal and discard the pairs. 
Note that for both row 1 and 2 in Table~\ref{tb:dpo_ablations}, we also discard the pairs with the difference of instruction following rewards (\selfrm) larger than $0.5$ (as mentioned in Section~\ref{subsubsec:factuality_aware_dpo}). 
Finally, for each response, we linearly combine the rewards of \selfrm{} (1--5 scale) and \factrm{} (0--1 scale) with the respective weight of 1 and 5 as a composite reward. 
For each instruction, we choose the responses with the maximum and minimum composite rewards as the positive and negative. 
As shown in Table~\ref{tb:dpo_ablations}, both data creation approaches increase the number of pairs in factuality preference data; however, it yields no obvious improvement in factuality but a bit degrade in instruction following capability (rows 2, 3 vs 1).

\begin{table}[t!]
	\caption{Effects of DPO training on response length.}
	\label{tb:impact_of_dpo_finetuning_on_generation_length}
	\centering
	 \resizebox{1\columnwidth}{!}{  
    \begin{tabular}{l|c|c|c|c}
\hline \hline
& \textbf{Alpaca Eval}& \textbf{Bio}& \textbf{Alpaca Fact}& \textbf{FAVA}\\
 \hline
 (1) \sft& \ \ 897&	1221& \ \ 969& \ \ 912\\
 \arrayrulecolor{lightgray}
 \hline
 \arrayrulecolor{black}
 (2) + \dpo& 1470&	1494& 1586& 1540\\
 (3) + \factonlydpo& 1160&	1166& 1192& 1104\\
 (4) + \fadpo& 1474&	1395& 1528& 1422\\

\arrayrulecolor{black}
\hline \hline
	\end{tabular}
	}
\end{table}

\paragraph{Impacts of DPO on Generation Length.} 
Table~\ref{tb:impact_of_dpo_finetuning_on_generation_length} lists the averaged length of models' responses for each dataset. 
We observe that DPO fine-tuned models tend to output lengthy responses than \sft{} except for \factonlydpo{} on Biography. 
This trend indicates that our instruction following reward model \selfrm{} guides LLMs to output more detailed and lengthy responses. 
In addition, we observe that although \fadpo{} outputs responses with similar length as \dpo{} on Alpaca Eval, \fadpo{} generates a bit shorter responses for the fact-based instructions in the other three datasets. 
This results show that our factuality-aware DPO training mainly impacts models' responses for fact-based instructions. 
The impact is mainly to reduce the output of false claims (see the numbers of correct and erroneous facts in rows 2 and 4 of Table~\ref{tb:main_dpo_results}).

\paragraph{Case Studies.} 
Figure~\ref{fig:case_studies} (in Appendix) showcases the generations of different models, \sft, \sft{} + \dpo{} and \fasft{} + \fadpo, on Alpaca Eval and Biography. 
Given the instruction, ``What are the names of some famous actors that started their careers on Broadway?'', \sft{} only lists some names of Broadway actors while DPO fine-tuned models generate detailed information for each listed Broadway actor. 
As for biography generations, we observe that given the instruction to generate a biography for a rare name entity, Marianne McAndrew, \sft{} + \dpo{} generates a detailed response but with many wrong facts while \sft{} and \fasft{} + \fadpo{} give relatively short responses. 
For the frequent entity, Ji Sung, all the models generate detailed and mostly correct responses. 
This qualitative analysis shows that \fasft{} + \fadpo{} tends to generate detailed responses for most instructions but for those instructions required tailed knowledge (e.g., rare entity) likely unknown to LLMs~\citep{mallen-etal-2023-trust}, it manages to reduce erroneous facts by giving less detailed responses, which is also observed by \citet{kang2024unfamiliar}.

\section{Conclusion}
In this paper, we present a study to enhance the factuality of large language models (LLMs). 
We first identify that the standard alignment approach, comprising SFT and RLAIF with DPO, may inadvertently encourage LLMs to produce more erroneous facts. 
Specifically, during the SFT stage, fine-tuning LLMs with high-quality human responses may introduce unfamiliar information, prompting LLMs to output unknown facts. 
Additionally, during the DPO stage, enhancing LLMs' ability to follow instructions may result in more detailed and lengthy responses but often leads to increased hallucination. 
To tackle the shortcomings of the standard alignment, we propose a factuality-aware alignment method, which includes factuality-aware SFT and DPO. 
Quantitative and qualitative analyses demonstrate that our factuality-aware alignment not only guides LLMs to generate detailed and helpful responses but also helps prevent the generation of false claims.

\section{Limitations}
While we have successfully integrated factuality into standard alignment procedure, our work only considers two alignment skill sets:\ instruction following (or helpfulness) and factuality. 
In practice, each instruction may require consideration of multiple and distinct alignment skill sets~~\citep{saha2023branchsolvemerge}. 
The method to optimize for these skill sets tailored to each query requires further study. 
In our experiments, we note that optimizing preferences solely for instruction following or factuality could potentially compromise the other. While our factuality-aware alignment demonstrated improvements in both aspects, it is uncertain whether there is a trade-off between the two aspects when integrating our approach to large-scale alignment~\citep{llama2}. 
Finally, as shown in Appendix, Figure~\ref{fig:fact_based_claim_check_cases}, not all the claims (or sentences) in a response require fact verification, a more accurate factuality reward model should take the factor into account. 
While our preliminary experiment, which removing non-fact-based sentences from the factuality reward modeling (Section~\ref{subsec:dicussions}), shows suboptimal performance, we believe that further study can bring more insights. 

\section*{Acknowledgements} 
We thank Bhargavi Paranjape for sharing fine-tuned Llama-2 7B for atomic fact decomposition and Jing Xu, Weizhe Yuan and Jason Weston for their helpful suggestions.

\bibliographystyle{acl_natbib}
\bibliography{paper.bib}

\clearpage

\appendix
\appendix
\section{Appendix}
\subsection{Biography Data Generation}
\label{appendix:biography_data_generation}
\paragraph{Entities for Training and Evaluation.} 
We use 500 diverse human entities to create training data for SFT and DPO; then, evaluate LLMs' generation factuality on another 183 human entities from \citet{factscore}.\footnote{\url{https://github.com/shmsw25/FActScore}}
Note that the human entities for training and evaluation are uniformly sampled from entities across diverse nationalities, professions, and rarities.
The instruction is generated with the format: Tell me a bio of \texttt{entity name}.

\paragraph{Creating Training Data for SFT.}
We randomly sample 5 human entities among the 500 entities for training and generate their biographies using Llama-2-Chat 70B as 5-shot demonstration.\footnote{\href{https://huggingface.co/meta-llama/Llama-2-70b-chat-hf}{meta-llama/Llama-2-70b-chat-hf}}
With the 5-shot demonstration, we use pre-trained Llama-2 7B to generate 10 biographies for each human entity from the remaining 495 ones.\footnote{\href{https://huggingface.co/meta-llama/Llama-2-7b}{meta-llama/Llama-2-7b}}  
We set temperature 0.7 and top-p 0.9 when generate multiple responses from LLMs in all our experiments. 
We use the created 4,950 name entity--biography pairs to fine-tune the pre-trained Llama-2 7B. 
As for generating training data with RAG, we prepend the top-10 passages from our retrieval system (detailed in Appendix~\ref{appendix:retrieval_models}) to each instruction and generate 10 biographies for each entity from RAG with 5-shot demonstrations. 
Note that we only prepend top-1 passage for each instruction in the demonstration. 

\paragraph{Creating Factuality Preference Pairs for DPO.} 
To construct factuality preference pairs, we first compute FActScore (FS) for all the 4,950 biographies previously created by \pt. 
Then, for each name entity, we compare the FS for all the possible 45 pairs from the 10 generated biographies and construct DPO pairs using the biography with a higher (lower) FS as a positive (negative). 
Note that we discard the pairs if they show tied FS. 

\subsection{Retrieval Models} 
\label{appendix:retrieval_models}
For each query, we retrieve top-$20$ candidate passages from Wikipedia using DRAGON+~\citep{dragon} and re-rank the candidates using a 12-layer cross-encoder\footnote{\href{https://huggingface.co/sentence-transformers/all-MiniLM-L12-v2}{sentence-transformers/all-MiniLM-L12-v2}}.
We use the Wikipedia version from the Dec. 20, 2021 dump released by \citet{atlas} in this work. 

\subsection{Alignment with Self Rewarding} 
\label{appendix:alignment_with_self_rewarding}
\paragraph{SFT.} 
At SFT stage, we fine-tune \pt{} on two seed datasets:\ (1) Instruction following training (IFT) data from \citet{li2024selfalignment}, consisting of 3200 instruction--response pairs created by humans from Open Assistant dataset~\citep[OASST;][]{oasst}, where we only use the first conversational turns in the English that are annotated rank 0;\footnote{\href{https://huggingface.co/datasets/OpenAssistant/oasst1}{OpenAssistant/oasst1}} (2) evaluation following training (EFT) data from \citet{selfrewarding}, the LLM-as-a-Judge data consists of 1630 samples, each of which contains instruction, human response and the corresponding score of 1-5 scale (with chain-of-though evaluation reasoning): $(x, y, r)$, where $(x, y)$ pairs are also selected from OASST other than training pairs and $r$ is created by the model fine-tuned only on IFT with manual filtering. 
The purpose of EFT is to enhance a LLM's capability as a reward model to judge the quality of a response in terms of relevance, coverage, usefulness, clarity and expertise. 
We refer readers to \citet{selfrewarding} for how EFT is created and filtered with minimum human efforts. 
The prompt template for LLM-as-a-Judge in EFT and an EFT training sample are shown in Appendix, Figure~\ref{fig:prompt_for_self_rewarding} and \ref{fig:eft_example}. 
We refer the baseline model fine-tuned on the IFT and EFT datasets as \sft. 

\paragraph{DPO for Instruction Following.} 
At the subsequent preference learning with DPO, following \citet{self-instruct}, we augment additional 20K instructions with Llama-2 70B chat model.\footnote{\href{https://huggingface.co/meta-llama/Llama-2-70b-chat-hf}{meta-llama/Llama-2-70b-chat-hf}}
For each augmented instruction $x$, we use \sft{} to generate 4 responses and evaluate how well the responses follow the instruction with score of 1--5 scale:\ $\text{\selfrm}(x, y_0) \cdots; \text{\selfrm}(x, y_3)$, where $y_0,\cdots,y_3 \in \text{\sft}(x)$ and \selfrm{} is the instruction following reward model. 
Note that, in self-rewarding~\citep{selfrewarding}, \selfrm{} is the same as \sft{} model. 
In addition, for each instruction--response pair, we use the same prompt in EFT seed data to sample the chain-of-thought evaluation three times and average the scores as the reward. 
Finally, for each instruction, we use the response with the highest (lowest) reward as the positive (negative) sample to form a preference pair for DPO training: $(x, y_{+}, y_{-})$. 
We discard the pair, if $\text{\selfrm}(x, y_+) = \text{\selfrm}(x, y_-)$.
In the DPO training, the model is initialized from \sft{} and the fine-tuned model is denoted \sft{} + \dpo.

\subsection{Factuality Reward Modeling} 
\label{appendix:factuality_reward_modeling}

\begin{figure*}[t]
    \centering
\includegraphics[width=\textwidth]
{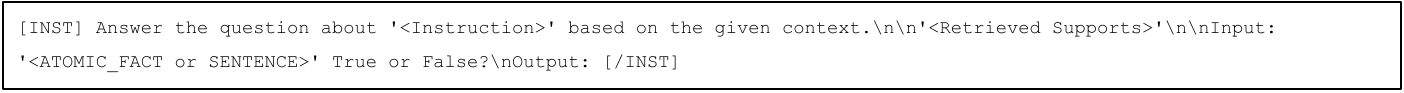}
   \caption{Prompt for fact check.} \label{fig:prompt_for_fact_check}
\end{figure*}

\paragraph{Factuality Reward Models.}
We build a reward model \factrm{} to measure the factuality of each response. 
The factuality reward model consists of two main modules. 
(1) fact decomposition:\ we first use \texttt{nltk.tokenize} to split a response into sentences; then, use our Llama-2 7B model fine-tuned on public datasets~\citep{liu-etal-2023-revisiting, chen-etal-2022-generating, malaviya2023expertqa} to conduct atomic fact decomposition for each sentence.\footnote{With few-shot demonstration, \sft{} is able to decompose a sentence into atomic facts with acceptable accuracy. Fine-tuning a Llama-2 7B is to reduce the inference time.} 
(2) Retrieval augmented claim verification:\ for each decomposed fact (or claim), we use the instruct Llama 7B fine-tuned on Super Natural Instructions~\citep{wang-etal-2022-super} to do fact check with the prompt shown in Figure~\ref{fig:prompt_for_fact_check}.\footnote{\href{https://huggingface.co/kalpeshk2011/instruct-llama-7b-wdiff}{instruct Llama 7B}} 
We append 10 retrieved supports (using the instruction as query) from our retrieval and re-ranking pipeline in Appendix~\ref{appendix:retrieval_models}.
Then, we compute the proportion of correct atomic facts in a response as a factuality reward.

\begin{table}[t!]
	\caption{A comparison of factuality reward models. $\tau$ denotes the correlation between human annotation.}
	\label{tb:factuality_rm_comparison}
	\centering
	 \resizebox{1\columnwidth}{!}{  
    \begin{tabular}{clccc}
\hline \hline
 & fact check model& \# sup.& fact unit& $\tau$\\
 \hline
 (1)& \multirow{2}{*}{Instruct Llama 7B}& 5& \multirow{2}{*}{atom.}& 0.32\\
 (2)& & 10& & 0.34\\
  \hline
 (3)& \multirow{2}{*}{\sft\ (Llama-2 70B)}& 5& \multirow{2}{*}{atom.}& 0.28\\
 (4)& & 10& & 0.31\\
 \hline
 (5)& \multirow{2}{*}{Instruct Llama 7B}& 5& \multirow{2}{*}{sent.}& 0.20\\
 (6)& & 10& & 0.25\\
\hline \hline
	\end{tabular}
	}
\end{table}

\paragraph{Quality of Factuality Reward Models.} 
We conduct ablation study on our factuality reward models. 
Specifically, we use our factuality reward models to detect the number of error facts in each instruction--response pair. 
We try different models for fact check using the prompt shown in Figure~\ref{fig:prompt_for_fact_check} with different numbers of retrieved supports. 
We use the LLMs' generated responses with human annotated hallucination provided by \citet{fava} to evaluate the quality of the factuality reward models.\footnote{\url{https://huggingface.co/datasets/fava-uw/fava-data/blob/main/annotations.json}}
Specifically, we rank the responses by numbers of errors detected and calculate the Kendall rank correlation ($\tau$) between the rank lists by our factuality reward models and humans. 
As shown in Table~\ref{tb:factuality_rm_comparison}, conducing fact check with more retrieved supports improves the accuracy of the factuality reward models (row 2 vs 1). 
In addition, our \sft{}, only fine-tuned on the IFT and EFT data, is capable of doing fact check, compared to Instruct Llama 7B fine-tuned on Super Natural Instructions~\citep{wang-etal-2022-super}. 
Finally, instead of computing the number of error facts from decomposed atomic facts, we conduct fact check directly for each sentence in a response and calculate the number of false sentences as error facts. 
However, the quality of the reward models shows significant decrease (rows 5,6 vs 1,2). 
We finally adopt row 2 as our factuality reward model.

\subsection{Training Details}
\label{appendix:training_details}
We fine-tune our models for 500 steps with a batch size of 32 and 64 on respective SFT and DPO stages. 
The learning rate and maximum sequence length is set to $1e-6$ (which decays to $1e-7$) and 2048, respectively. 
At SFT stage, we mix the IFT and EFT while at DPO stage, we set $\beta = 0.1$ and  uniformly sample between self rewarding $(x, y_{+}, y_{-})$ and factuality reward $(x, y_{\text{\tiny true}}, y_{\text{\tiny false}})$ preference data. 
Note that \sft{} (\fasft) + \dpo{} meaning that we use \sft{} (\fasft) to create preference data, serve as instruction following reward model \selfrm and as the initialization of DPO. 
The data used to fine-tune different variants are listed in Table~\ref{tb:training_data_statistics}. 

\begin{table}[t]
	\caption{Training data statistics for different variants. IF.~and Fact.~denote instruction following $(x, y_{+}, y_{-})$ and factuality $(x \in X^{\text{\tiny fact}}, y_{\text{\tiny true}}, y_{\text{\tiny false}})$ preference data, where $X^{\text{\tiny fact}}$ denotes the set of fact-based instructions.}
	\label{tb:training_data_statistics}
	\centering
	 \resizebox{1\columnwidth}{!}{  
    \begin{tabular}{l|cc|cc}
\hline \hline
& \multicolumn{2}{c}{\textbf{Seed IFT (\# of Inst.)}}& \multicolumn{2}{c}{\textbf{Preference (\# of pairs)}}\\
  \cmidrule(rl){2-3} \cmidrule(rl){4-5}
model variant&$x \notin X^{\text{\tiny fact}}$ & $x \in X^{\text{\tiny fact}}$&IF. & Fact.\\
\hline
\sft\ + \dpo& \multirow{3}{*}{2,187}& \multirow{3}{*}{1,013}& 18,454& - \\ 
\sft\ + \factonlydpo& & & -& 3,315\\ 
\sft\ + \fadpo& & & 18,454& 3,315\\ 
\hline
\fasft\ + \dpo& \multirow{2}{*}{2,187}& \multirow{2}{*}{1,013}& 18,603& - \\ 
\fasft\ + \fadpo& & & 18,603& 4,211 \\ 
\hline \hline
	\end{tabular}
	}
\end{table}

\label{appendix:eft_creation}
\begin{figure*}[t]
    \centering
\includegraphics[width=\textwidth]
{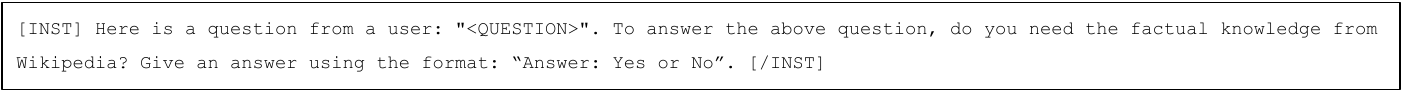}
\caption{Prompt to check whether an instruction is fact-based.} 
    \label{fig:prompt_for_instruction_classifier}
\end{figure*}

\begin{figure*}[t]
    \centering
\includegraphics[width=\textwidth]
{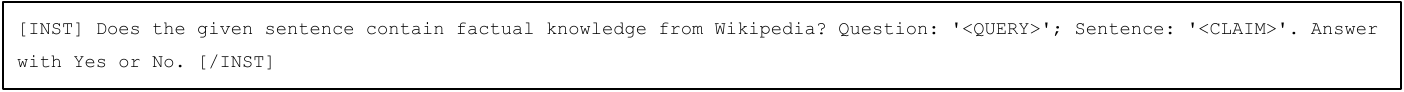}
\caption{Prompt to check whether a claim is fact-based.} 
    \label{fig:prompt_for_claim_classifier}
\end{figure*}

\begin{figure*}[t]
    \centering
\includegraphics[width=\textwidth]
{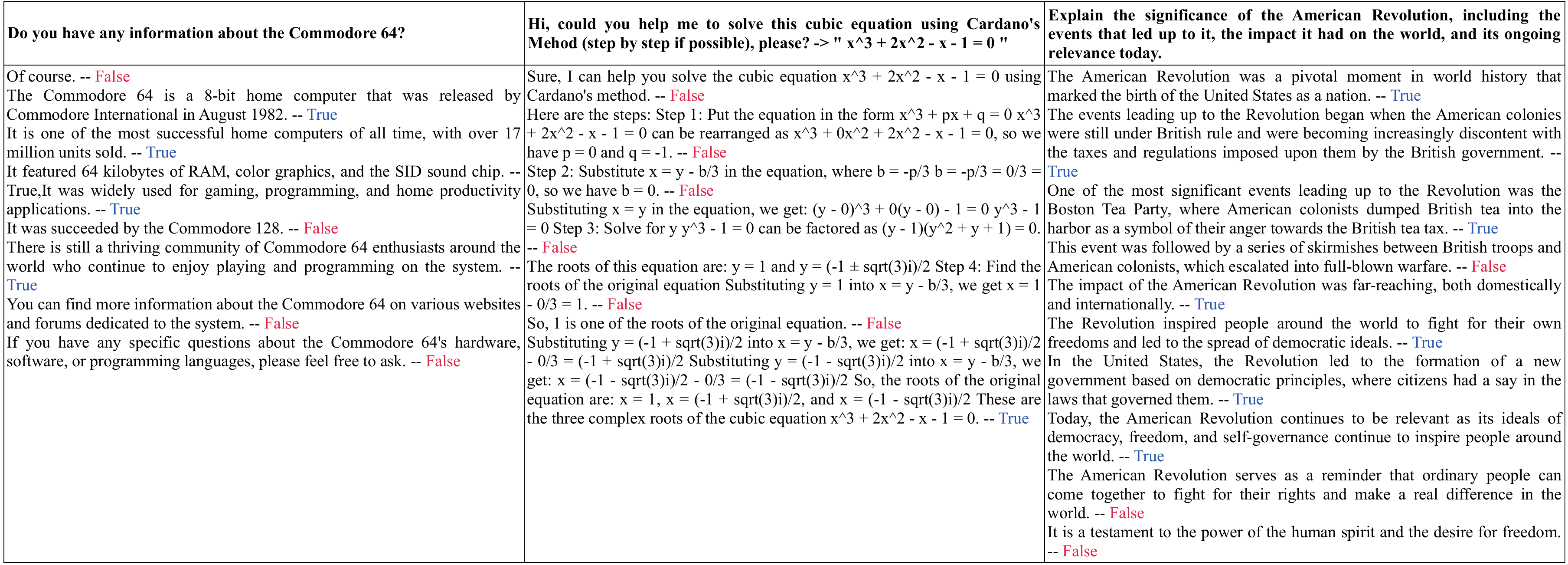}
\caption{The results of whether a sentence is fact-based or not classified by \sft{} with prompt in Figure~\ref{fig:prompt_for_claim_classifier}.} 
    \label{fig:fact_based_claim_check_cases}
\end{figure*}

\begin{figure*}[t]
    \centering
\includegraphics[width=\textwidth]
{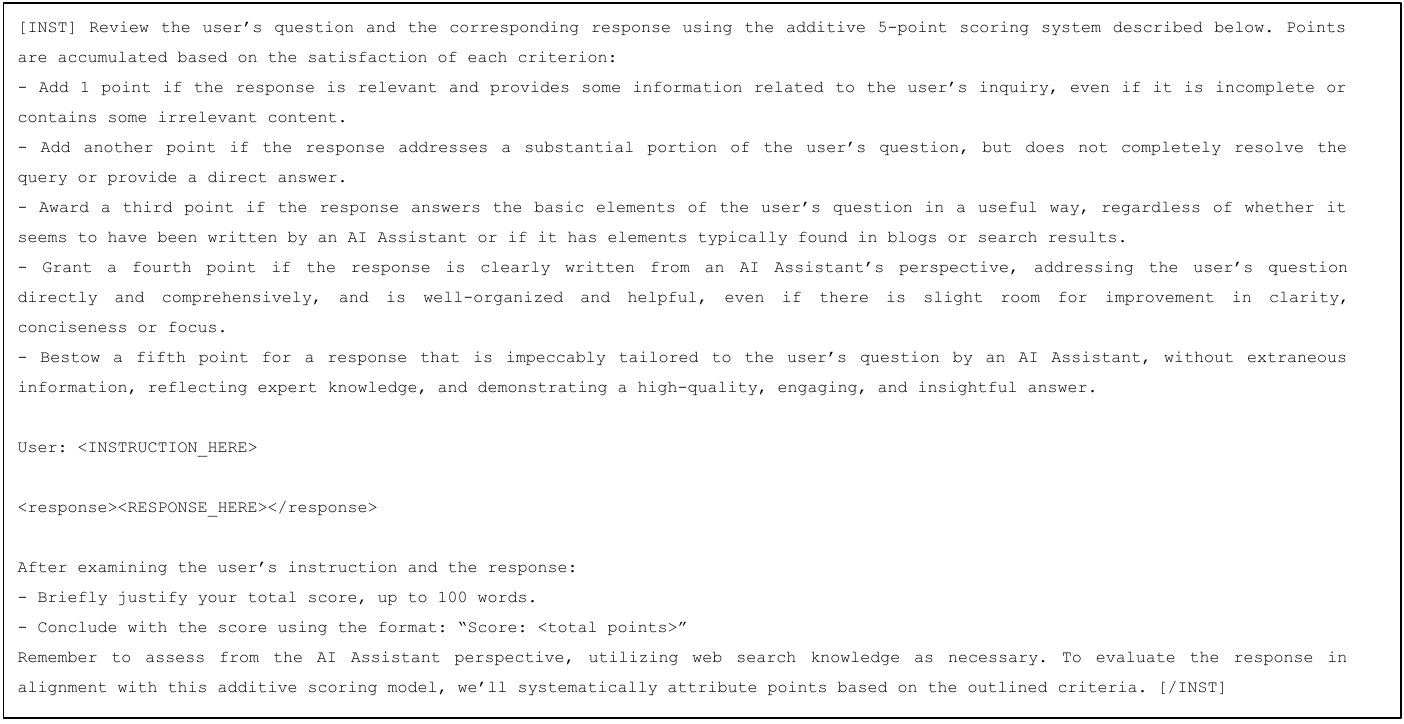}
   \caption{Prompt to evaluate models' instruction following capability from \citet{selfrewarding}.} \label{fig:prompt_for_self_rewarding}
\end{figure*}

\begin{figure*}[t]
    \centering
\includegraphics[width=\textwidth]
{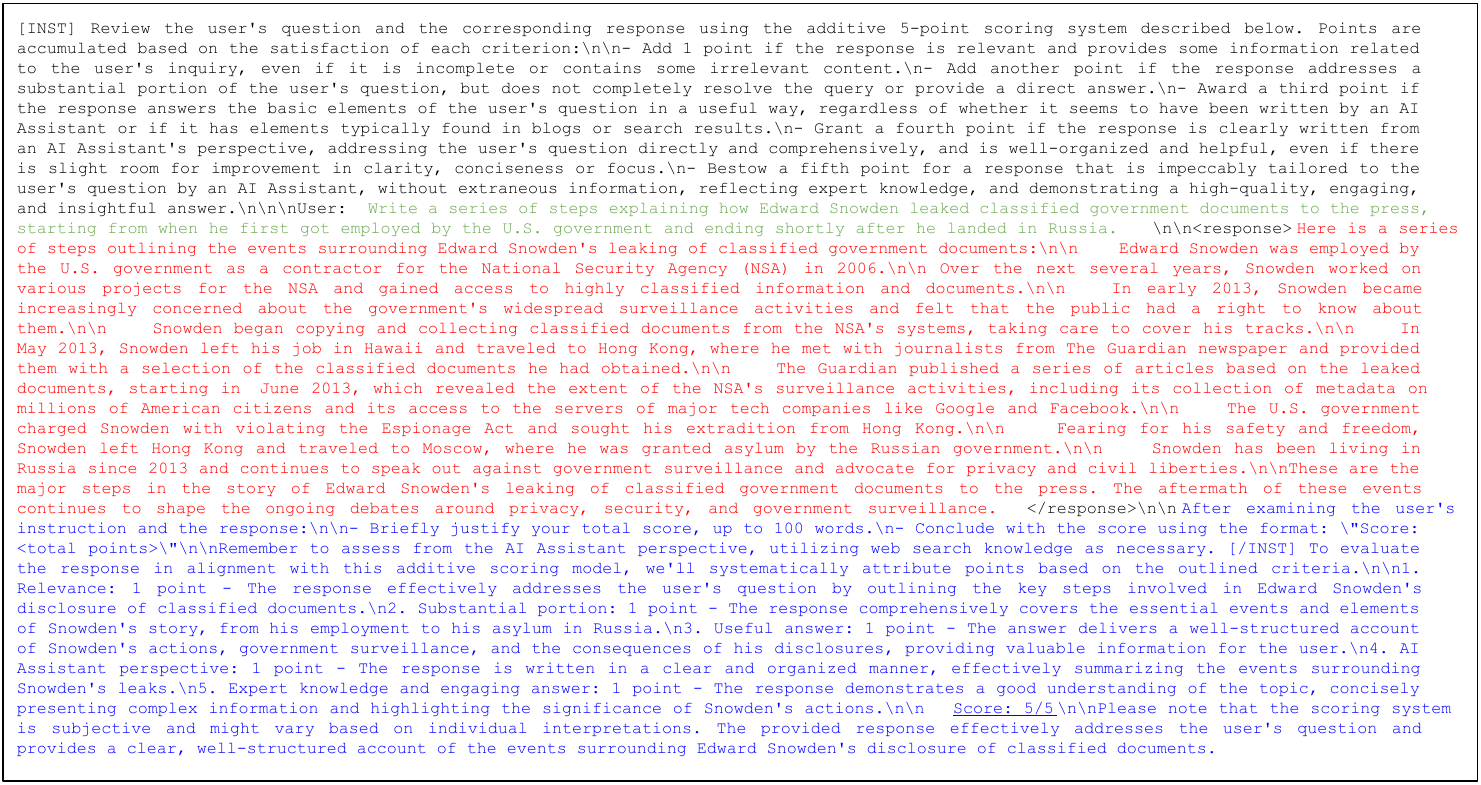}
   \caption{An example of EFT data. The texts with the colors of green, red and blue are the instruction, response and the LLM-as-a-judge results (explanation and score), respectively} 
    \label{fig:eft_example}
\end{figure*}

\begin{figure*}[t]
    \centering
\includegraphics[width=\textwidth]
{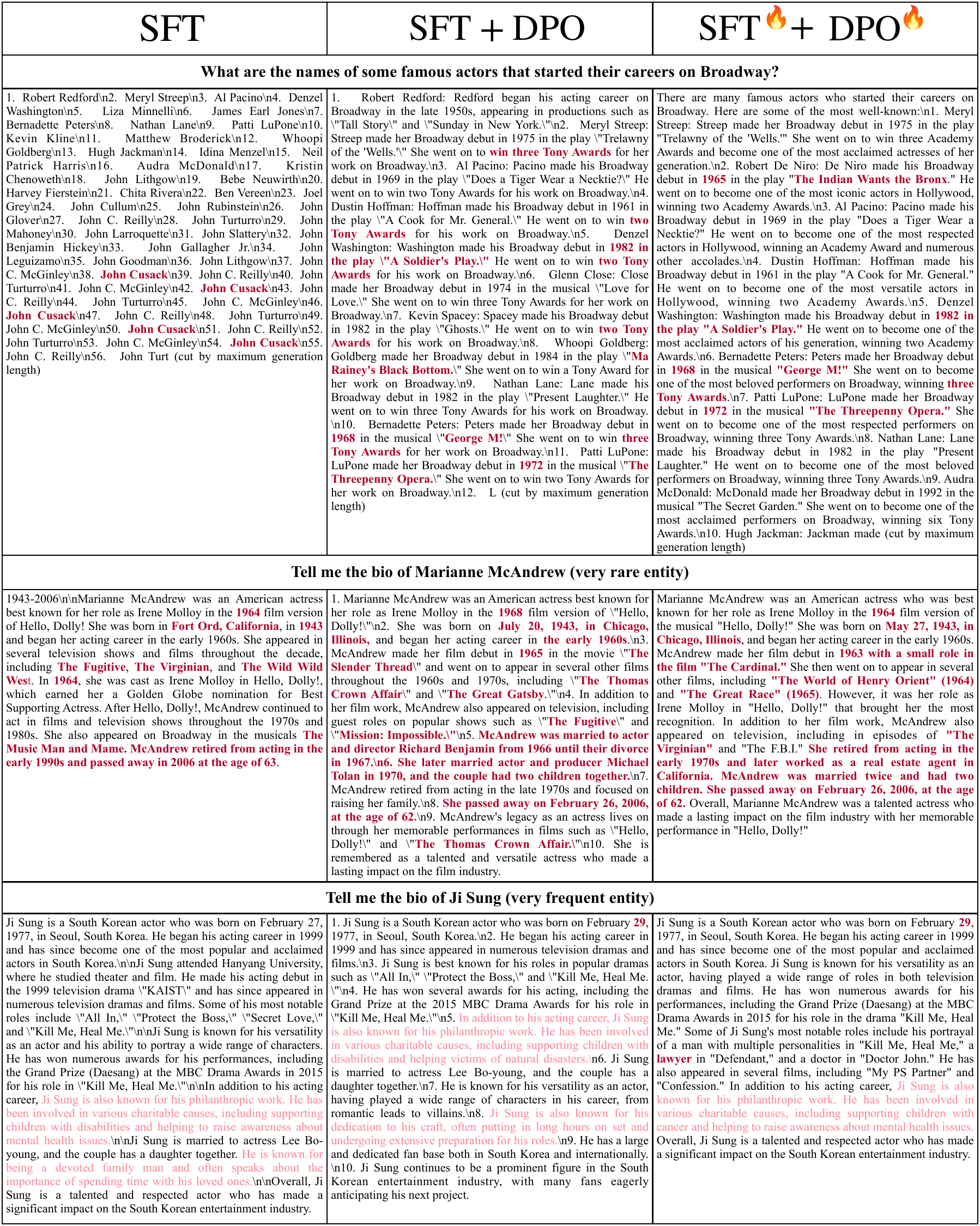}
\caption{Generation comparisons for instructions from Alpaca Eval and Biography (very rare and frequent entities). Determined through manual verification using Google search, red denotes incorrect identified facts while pink indicates unverified facts; e.g., we cannot search relevant pages about Ji Sung's involvement in charitable causes but also cannot dismiss the possibility of his contributions. 
Note that the popularity of an entity is defined by its occurrence and page views in Wikipedia, which are provided by \citet{factscore}.} 
    \label{fig:case_studies}
\end{figure*}

\end{document}